\documentclass{article}
\usepackage{amsmath}
\usepackage{amsfonts}

\usepackage{graphicx} % more modern
\usepackage{subfigure} 
\usepackage{lipsum}
\usepackage{color}
\usepackage{comment} 
\excludecomment{versionB}

\usepackage{natbib}

\usepackage{algorithm}
\usepackage{algorithmic}

\usepackage{hyperref}

\usepackage[accepted]{icml2013} 
\icmltitlerunning{Learning Program Embeddings to Propagate Feedback on Student Code}

\usepackage{balance}  % to better equalize the last page
\usepackage{graphics} % for EPS, load graphicx instead
\usepackage{times}    % comment if you want LaTeX's default font
\usepackage{url}      % llt: nicely formatted URLs
\usepackage{booktabs} 
\usepackage{mdframed}
\usepackage{booktabs}
\usepackage{lipsum}
\usepackage{xcolor}
\usepackage{mdwlist}
\usepackage{dblfloatfix}
\usepackage[english]{babel}
\usepackage{blindtext}

\usepackage{soul}
\usepackage{tikz}

\usetikzlibrary{calc}

\usepackage{etex}

\newcommand{\ra}[1]{\renewcommand{\arraystretch}{#1}}
\newcommand\tabhead[1]{\small\textbf{#1}}

\newcommand{\Jon}[1]{\textcolor{blue}{#1}}

\newcommand{\reals}{\mathbb{R}}
\begin{document} 

\twocolumn[
\icmltitle{Learning Program Embeddings to Propagate Feedback on Student Code}
%Idea 0: Learning Program Embeddings to Automate Program Assessment
%Idea 1: Program Embeddings That Enable Machine Learning for CS education
%Idea 2: Learning representations of programs with applications to CS education
%emphaiss is on embeding a collection of programs of related programs.
%Feedback propagations in programs by embedding into a shared space.
%Providing program feedback by embedding related programs into a shared space.

% It is OKAY to include author information, even for blind
% submissions: the style file will automatically remove it for you
% unless you've provided the [accepted] option to the icml2013
% package.
\icmlauthor{Chris Piech}{piech@cs.stanford.edu}
\icmlauthor{Jonathan Huang}{jonathanhuang@google.com}
\icmlauthor{Andy Nguyen}{tanonev@cs.stanford.edu}
\icmlauthor{Mike Phulsuksombati}{mikep15@cs.stanford.edu}
\icmlauthor{Mehran Sahami}{sahami@cs.stanford.edu}
\icmlauthor{Leonidas Guibas}{guibas@cs.stanford.edu}

%\icmladdress{353 Serra Mall, Stanford, CA 94305 USA}

% You may provide any keywords that you 
% find helpful for describing your paper; these are used to populate 
% the "keywords" metadata in the PDF but will not be shown in the document
\icmlkeywords{boring formatting information, machine learning, ICML}

\vskip 0.3in
]

\begin{abstract} 
Providing feedback, both assessing final work and giving hints to stuck students, is difficult for open-ended assignments in massive online
classes which can range from thousands to millions of students. We introduce a neural network method to encode programs as a linear mapping from an embedded precondition space to an embedded postcondition space and propose an algorithm for feedback at scale using these linear maps as features. We apply our algorithm to assessments from the Code.org Hour of Code and Stanford University's CS1 course, where we propagate human comments on student assignments to orders of magnitude more submissions.
\end{abstract} 
\vspace{-7mm}
\section{Introduction}\vspace{-1mm}
Online computer science courses can be massive
with numbers ranging from thousands to even millions of students.
%
%Online computer science courses have thousands, and sometimes even millions of students. 
%Though the delivery of information 
Though technology has increased our ability to provide 
content
%personalized feedback 
to students at scale, assessing and providing feedback 
(both for final work and partial solutions) remains difficult.
%giving hints to stuck students, has not. 
Currently, giving personalized feedback, a staple of quality education, is costly for small, in-person classrooms and prohibitively expensive for massive classes. 
Autonomously providing feedback is therefore 
a central challenge for at scale computer science education.

It can be difficult to apply machine learning directly to
data in the form of programs.  Program representations
such as the \emph{Abstract Syntax Tree (AST)} are not directly conducive to standard statistical methods and the edit distance metric between such trees 
%, a standard component in methods for deciding when programs are similar enough to share the same feedback,
are not discriminative enough to be used to share feedback accurately since programs with similar ASTs can behave quite differently and require different comments. Moreover, though unit tests are a useful way to test if final solutions are correct they are not well suited for giving help to students with an intermediate solution and they are not able to give feedback on stylistic elements.

%\Jon{this is very abrupt.}
%Student programs have natural patterns and shared structure. 
%However, utilizing the underlying commonalities is difficult. The Abstract Syntax Tree (AST) representations of the programs, while expressive, are not directly conducive to statistical methods. Moreover, the edit distance metric between such trees, a staple of state-of-the-art methods for propagating information amongst programs, requires quadratic number of computations with respect to the size of the corpus and is not discriminative enough to be used to share feedback: similar syntax trees, can behave quite differently, and require different comments. 

\begin{figure}[t]
\centering
\includegraphics[width=.95\columnwidth]{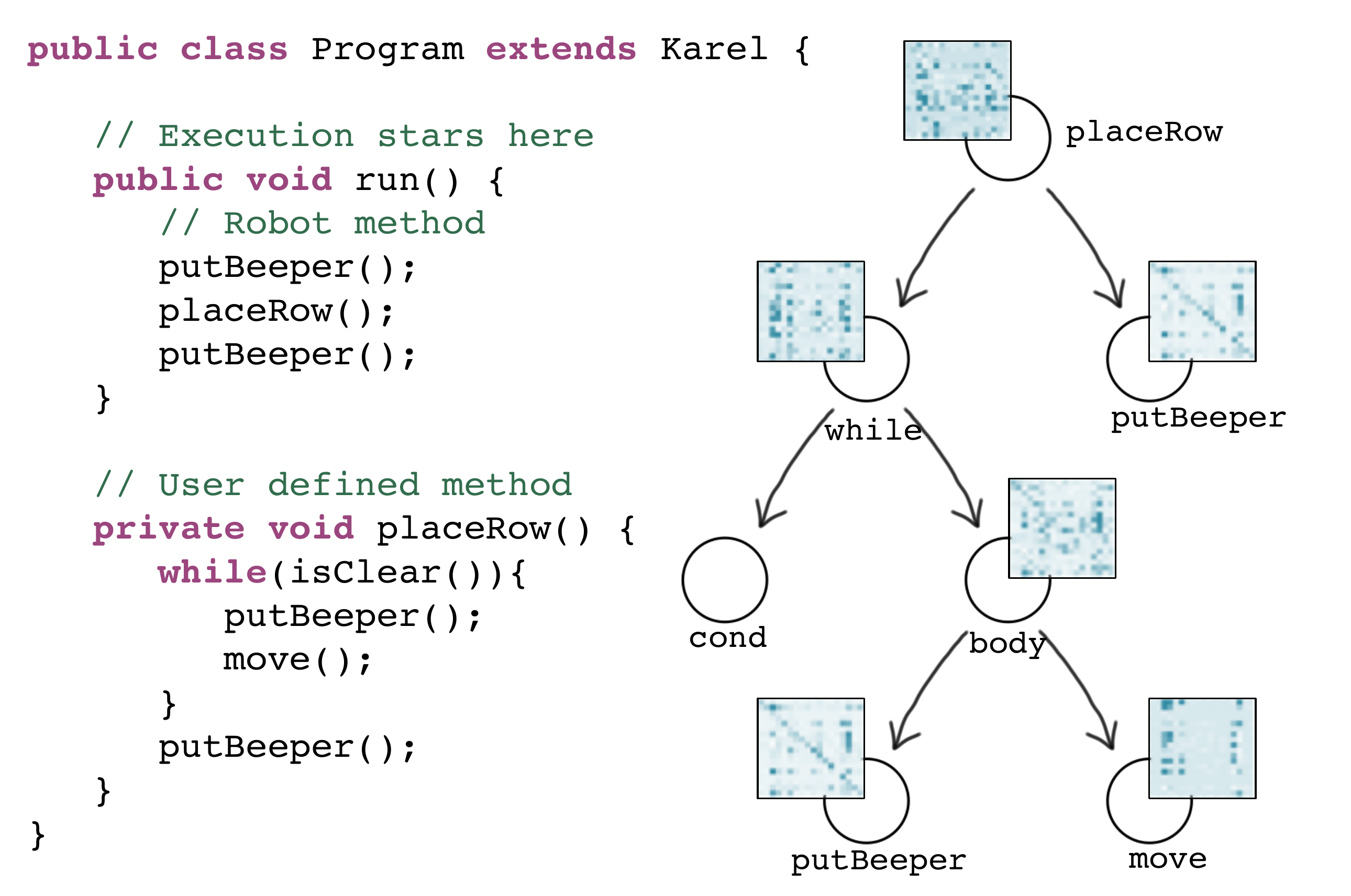}\vspace{-2mm}
\caption{\footnotesize We learn matrices which capture functionality. Left: a student partial solution. Right: learned matrices for the syntax trees rooted at each node of placeRow.}\vspace{-2mm}
\label{fig:splash}
\end{figure}

There are two major goals of our paper.  The 
first is to automatically learn a feature embedding of 
student submitted programs
that captures functional and stylistic elements and can be easily
used in typical supervised machine learning systems.
The second is to use these features to learn
how to give automatic feedback to students. Inspired by recent successes of deep learning for learning features 
in other domains like NLP and vision, 
we formulate a novel neural network architecture 
that allows us to jointly optimize an embedding of programs and memory-state in a feature space.
%We propose the use of neural networks to jointly learn an embedding of states and programs that best captures these pre-condition, program, post-condition triples. 
See Figure~\ref{fig:splash} for an example program and corresponding matrix embeddings.

%A unique property of programs is that they are 
To gather data, we exploit the fact that programs are 
executable --- that we can evaluate any piece of code on 
an arbitrary input (i.e., the precondition), and observe the state  
after, (the postcondition). For a program and its 
constituent parts we can thus collect arbitrarily many such precondition/postcondition mappings. This %deluge of data 
data provides the training set from which we can learn a shared representation for programs. 
To evaluate our program embeddings we test our ability to amplify teacher feedback. We use real student data from the Code.org Hour of Code which has been attempted by over 27 million learners
%and has been taught in over 90 thousand classrooms, 
making it, to the best of our knowledge,
the largest online course to-date. 
We then show how the same approach can be used for submissions in Stanford University's Programming Methodologies course which has thousands of students and assignments that are substantially more complex. The programs we analyze are written in a Turing-complete language but do not allow for user-defined variables.

Our main contributions are as follows. First, we present a 
method for computing features of code that capture both functional
and stylistic elements.  Our model works by simultaneously embedding precondition and postcondition spaces of 
a set of programs into a feature space where programs can be viewed as linear maps on this space.
Second, we show how our code features can be useful for automatically
propagating instructor feedback to students in a massive course.
Finally, we demonstrate the effectiveness of our methods
on large scale datasets.
%model 
%which allows us to simultaneously embed input and output spaces of 
%a collection of programs into a Euclidean space where programs can be viewed as linear map on this space. Second, we demonstrate the effectiveness of this model applied to two tasks: that of predicting the results of running a program on a given precondition and of propagating teacher feedback to a large number of students. 
%Finally, we explore the relationship between program complexity and the performance of machine learning tasks using our program embeddings.
Learning embeddings of programs is fertile ground for machine learning research and if such embeddings can be useful for the propagation of teacher feedback this line of investigation will have a sizable impact on the future of computer science education.

\section{Related Work}\label{sec:related}
The advent of massive online computer science courses 
has made the problem of automated reasoning with large code
collections an important problem.
There have been a number of recent papers~\cite{huangetal13,basu2013powergrading,nguyen14,brooks2014divide,lan2015mathematical, piech2015} on using large
homework submission datasets to improve student feedback. The volume of work speaks to the importance of this problem. Despite the research efforts, however, providing quality feedback at scale remains an open problem. 

A central challenge that a number of papers
 address is that of measuring similarity between 
source code.
Some authors have done this without an explicit 
featurization of the code ---
for example, the \emph{AST edit distance} has been a popular choice~\cite{huangetal13,rogers2014aces}.
\cite{mokbel2013domain} explicitly 
hand engineered a small collection of features on ASTs
that are meant to be domain-independent. 

To incorporate functionality,
\cite{nguyen14} proposed a method that discovers program modifications that do not appear to change the semantic meaning of code. 
The embedded representations of programs used in this paper also capture semantic similarities and are more amenable to prediction tasks such as propagating feedback. 
We ran feedback propagation on student data using methods from Nguyen et al and observe that embeddings enabled notable improvement (see section 6.3).
%While this approach was useful,
%determining absolute equivalence is an unsolvable problem and it quickly becomes hard to decide when we have enough evidence to label code as being the same. The methods in this paper allow for different degrees of similarity and a more direct means to use program similarities to propagate feedback. %it is not fully automated, which makes its application to new problems difficult, and is unable to 
%and while it helps reduce the complexity of a corpus of student programs the extent to which it can be used to propagate functional feedback is constrained by the large combinatorial number of student solutions and such a method is unable to 
%disseminate stylistic comments.
%Our method, by contrast, allows for more general propagation of comments and can be inserted into other machine learning systems.

Embedding programs has many crossovers with embedding natural language artifacts, given the similarity between the AST representation and parse trees. Our models are related to recent work from the NLP and deep learning communities on recursive neural networks, particularly
for modeling semantics in sentences or symbolic expressions~\cite{socher2013recursive,socher2011semi,zaremba2014learningb,bowman2013can}.
%illustrate the effectiveness of Recursive Neural Networks (RNNs) and Recursive Neural Tensor Networks. 

Finally, representing a potentially 
complicated function (which in our case is a program) as
a linear operator acting on a nonlinear feature space has also
been explored in different communities.
The computer graphics community
have represented pairings of nonlinear geometric
shapes as linear maps between shape features, called \emph{functional maps}~\cite{ovsjanikov2012functional,ovsjanikov2013analysis}.
From the kernel methods literature, there has also been recent
work on representations of conditional probability
 distributions as operators on a Hilbert space~\cite{song2013kernel,song2009hilbert}. 
From this point of view, our work is novel in that it
focuses on the joint optimization of feature embeddings
together with a collection of maps
so that the maps simultaneously ``look linear''
with respect to the feature space.

\vspace{-2mm}
\section{Embedding Hoare Triples}\label{sec:embedding}
\vspace{-1mm}
\begin{figure*}[t]
\centering
\includegraphics[width=1.0\textwidth]{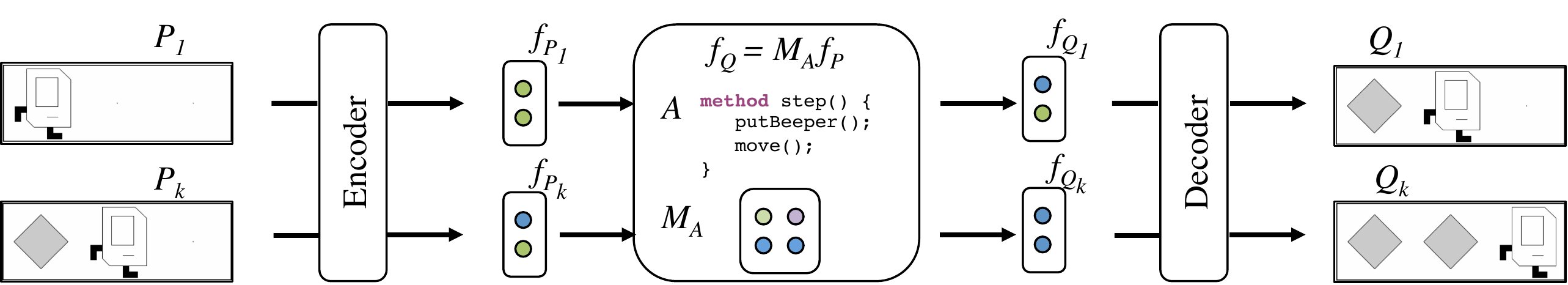}\vspace{-2mm}
\caption{\footnotesize
Diagram of the model for a program $A$ implementing a simple ``step forward'' behavior
in a small 1-dimensional gridworld. Two of the $k$ Hoare triples that correspond with $A$ are shown. Typical worlds are larger and programs are more complex.
}
\label{fig:hoare}
\vspace{-2mm}
\end{figure*}
Our core problem is to represent a program as a point in a fixed-dimension real-valued space that can then be used directly as input for typical supervised learning algorithms.

While there are many dimensions that ``characterize'' a program including
aspects such as style or time/space complexity, we begin by 
first focussing on capturing the most basic aspect of a program --- its function.  While capturing the function of the program ignores
aspects that can be useful in application (such as giving
stylistic feedback in CS education), we discuss in later
sections how elements of style can be recaptured by modeling the
function of subprograms that correspond to each subtree of an AST.
Given a program $A$ (where we consider a program to generally be any executable code whether a full submission or a subtree of a submission), and a precondition $P$, we thus would
like to learn features of $A$ that are useful for predicting the outcome of running $A$ when $P$ holds.  In other words, we want to 
predict a postcondition $Q$ out of some space of possible postconditions. Without loss of generality we let $P$ and $Q$ be 
real-valued vectors encapsulating the ``state'' of the program (i.e., the values of all program variables)
at a particular time. For example, in a grid world, this vector would contain the location of the agent, the direction the agent is facing, the status of the board and whether the program has crashed. Figure \ref{fig:hoare} visualizes two preconditions, and the corresponding postconditions for a simple program. 

We propose to learn program features using a training set of $(P, A, Q)$-triples --- so-called \emph{Hoare triples}~\cite{hoare1969axiomatic}
obtained via historical runs of a collection of programs on a collection of preconditions.
%The dimensions
%of $P$ and $Q$ are assumed to be fixed across all instances in a training set.
We discuss the process by which such a dataset can be obtained  in Section~\ref{sec:data}. 
The main approach that we espouse in this paper is to simultaneously find an embedding of states and programs into feature space where pre and postconditions are points in this space and programs are mappings between them.

The simple way that we propose to relate preconditions to postconditions
is through a linear transformation.  Explicitly, given a
$(P, A, Q)$-triple, if $f_P$ and $f_Q$ are $m$-dimensional nonlinear feature
representations of the pre and postconditions $P$ and $Q$, respectively,
then we relate the embeddings via the equation
\begin{equation}\label{eqn:linearmap}
f_Q = M_A \cdot f_P.   
\end{equation}
We then take the $m\times m$ matrix of coefficients $M_A$
as our feature representation of the program $A$ and refer to it as
the \emph{program embedding matrix}. 
We will want to learn the mapping into feature space $f$ 
as well as the linear map $M_A$ such that 
this equality holds for all observed triples and can
generalize to predict postcondition $Q$ given $P$ and $A$.

At first blush, this linear relationship may seem too limiting
as programs are not linear nor continuous in general.  By learning a nonlinear embedding function $f$ for the pre and postcondition spaces,
however, we can capture a rich family of nonlinear relationships
much in the same way that kernel methods allow for nonlinear decision boundaries.

As described so far, there remain a number of modeling choices to be made. In the following, we elaborate further on how we model the
feature embeddings $f_P$, and $f_Q$ of the pre and postconditions, and how to model the program embedding matrix $M_A$.

\subsection{Neural network encoding and decoding of states}
We assume that preconditions have some base encoding as a $d$-dimensional vector, which we refer to as $P$.
For example, in image processing courses, the state space could simply be
the pixel encoding of an image, whereas
in the discrete gridworld-type programming problems that we use in our experiments, we might choose to encode the $(x,y)$-coordinate
and discretized heading of a robot using a concatenation of 
one-hot encodings. 
Similarly, we assume that there is a base encoding $Q$ of the postcondition.

We will focus our exposition in the remainder of our paper
on the case where the precondition space 
and postcondition spaces share a common base encoding.  This is particularly
appropriate to our experimental setting in which both the preconditions
and postconditions are representations of a gridworld.  In this case, we can use the same decoder parameters 
(i.e., $W^{dec}$ and $b^{dec}$) to decode both from precondition space and postcondition space
 --- a fact that we will exploit in the following section.

Inspired by nonlinear autoencoders, we parameterize a mapping, called
the \emph{encoder} from 
precondition $P$ to a nonlinear $m$-dimensional
feature representation $f_P$.
As with traditional autoencoders, we use an affine mapping composed
with an elementwise nonlinearity:  
%Using $\tanh$ nonlinearities (as is done
%in our experiments), we would thus have:
\begin{equation}\label{eqn:encodeprecondition}
f_P =  \phi(W^{enc}\cdot P + b^{enc}),
\end{equation}
where $W^{enc}\in\reals^{m\times d}$, $b^{enc}\in \reals^m$,
and $\phi$ is an elementwise nonlinear function (such as $\tanh$).
At this point, we can use the representation
$f_P$ to decode or reconstruct the original precondition as 
a traditional autoencoder would do using:
\begin{equation}\label{eqn:decodeprecondition}
\hat{P} =  \psi (W^{dec}\cdot f_P + b^{dec}),
\end{equation}
where $W^{dec}\in\reals^{d\times m}$, $b^{dec}\in\reals^{d}$,
and $\psi$ is some (potentially different) elementwise 
nonlinear function.
Moreover, we can push the precondition
embedding $f_P$ through Equation~\ref{eqn:linearmap}, and decode the
postcondition embedding $f_Q=M_A\cdot f_P$.  This mapping which reconstructs
the postcondition $Q$, the \emph{decoder}, takes the form:\vspace{-5mm}

{\footnotesize
\begin{align}
\hat{Q} &= \psi (W^{dec}\cdot f_Q + b^{dec}), \label{eqn:decodepostcondition}\\
    &= \psi (W^{dec}\cdot M_A\cdot f_P + b^{dec}).
\end{align}\vspace{-5mm}
}

%Rather than being interpreted as a strict reconstruction of $Q$,
%the softmax function allows us to interpret $\hat{Q}$ as a probability
%distribution over the space of possible postconditions.
Figure~\ref{fig:hoare} diagrams our model on a simple program.
Note that it is possible to swap in alternative feature representations.
We have experimented with using a deep, stacked autoencoder however our results have
not shown these to help much in the context of our datasets.

\subsection{Nonparametric model of program embedding}\label{sec:nonparametric}
%Thus far, we have discussed encoder and decoder parameters for our model.
%What remains is a model for the program embedding matrix. 
To encode the program embedding matrix, we propose a simple
nonparametric model in which each program in the training set is associated
with its own embedding matrix.
Specifically, if the collection of unique programs
is $\{A_1,\dots,A_m\}$, then for each $A_i$, we will associate a matrix $M_i$.
The entire parameter set for our nonparametric
matrix model (henceforth abbreviated \emph{NPM}) is thus:
$\Theta=\{W^{dec}, W^{enc}, b^{enc}, b^{dec}\}\cup\{M_i\,:\,i=1,\dots,m\}$.

To learn the parameters, we minimize a sum of three terms: (1) a prediction loss
$\ell^{pred}$ which quantifies how well we can predict postcondition of a program 
given a precondition, (2) an autoencoding loss $\ell^{auto}$ which quantifies how good 
the encoder and decoder parameters are for reconstructing given preconditions, and (3) a regularization term $\mathcal{R}$.
Formally, given training triples $\{(P_i, A_i, Q_i)\}_{i=1}^n$,
we can minimize the following objective function:\vspace{-3mm}

{\footnotesize
\begin{align}
\begin{split}
L(\Theta) &=  \frac{1}{n} \sum_{i=1}^n 
                \ell^{pred}(Q_i, \hat{Q}_i(P_i, A_i; \Theta)) \\
        %\ell^{pred}(\Theta; Q_i, \hat{Q}_i, Pi, A_i)   \\
    &\;\;+  \frac{1}{n} \sum_{i=1}^n 
                    \ell^{auto}(P_i, \hat{P}_i(P_i, \Theta)) 
        %    \ell^{auto}(\Theta; Q_i, \hat{Q}_i, Pi, A_i) 
    + \frac{\lambda}{2}\mathcal{R}(\Theta), \label{eqn:objective}
  \end{split}
\end{align}
}\vspace{-3mm}

where
$\mathcal{R}$ is a regularization term on the parameters,
and $\lambda$ a regularization parameter.
In our experiments, we use $\mathcal{R}$ to penalize the sum of the $L_2$ norms of the
weight matrices (excluding the bias terms $b^{enc}$ and $b^{dec}$).

Any differentiable loss can conceptually be used for $\ell^{pred}$ and
$\ell^{auto}$.  For example,
when the top level predictions, $\hat{P}$ or $\hat{Q}$,
can be interpreted as probabilities (e.g., when $\phi$ is
the Softmax function), we use a cross-entropy loss function.
%\vspace{-1mm}
%{\footnotesize
%\begin{align*}
%\ell^{pred} &= \frac{1}{n} \sum_{i=1}^n \log(p(Q_i \,|\, %\hat{Q}_i(P_i, A_i; \Theta))), \\
%\ell^{auto} &= \frac{1}{n} \sum_{i=1}^n \log (p(P_i \,|\, %\hat{P}_i(P_i; \Theta) )),
%\end{align*}
%}

Informally speaking, one can think of our optimization problem (Equation~\ref{eqn:objective}) as trying to
find a good \emph{shared} representation of the state space ---
shared in the sense that 
even though programs are clearly not linear maps over the
original state space, the hope is that we can discover some 
nonlinear encoding of the pre and postconditions such that most
programs simultaneously 
``look'' linear in this new projected feature space.
As we empirically show in Section~\ref{sec:experiments}, such a representation
is indeed discoverable.

We run joint optimization using minibatch stochastic gradient 
descent without momentum, 
using ordinary backpropagation to calculate the gradient.
We use random search~\cite{bergstra2012random}
to optimize over hyperparameters (e.g, regularization parameters,
matrix dimensions, and minibatch size).
Learning rates are set using Adagrad~\cite{duchi2011adaptive}. We seed our parameters using a ``smart'' initialization 
in which we first learn an
autoencoder on the state space, and perform a vector-valued ridge
regression for each unique program
to extract a matrix mapping the features of the precondition
to the features of the postcondition.  The encoder and decoder parameters and the program matrices are then jointly optimized.

%\Jon{We initialize gradient descent
%by randomizing all parameters.  Does this belong here or at the top of %the paragraph?
%And just for parametric model.}

%The idea here is that we would like to learn a representation of the %pre and post conditions
%that's good for representing multiple programs
% simultaneously as linear maps.
% We can discuss how for a neural network autoencoder
% this is always possible (in terms of representational
% power) for a single program.
%I.E. We are finding shared structure amongst states
%and shared structure amongst the transformations.

%Thusly formulated, our model bears some resemblance to traditional
%autoencoders, particularly when the precondition and postcondition
%spaces are the same.

%Introduce the general idea of embedding a program
%as a fixed size representation.
%Need to account for inputs to capture
%functionality, so propose matrix representation.
%
%Introduce our model: Hoare triple embedding. Capture all Huare triples for a given piece of code.
%Explain encoder, decoder.  Discuss loss functions ---
%emphasize the point that we are training the 
%representation of the pre/post conditions with the
%representation of the transformation.

%Discuss the fact that when precondition space and
%postcondition space are the same (which is sometimes
%the case in educational programs), we can enforce loss
%at two points

\begin{versionB}

\subsection{Parametric baseline}
As a baseline, we compare the nonparametric model of programs to a more constrained parametric forumulation based on recursive neural networks
(RNNs).
Instead of learning one parameter matrix for each unique program,
the RNN model parameterizes the matrix $M_A$ using a neural network architecture which follows the abstract syntax tree of the program
(similar to the way in which RNN architectures might take the form of a parse tree in an NLP 
setting~\cite{socher2013recursive}). 

%The delta from Richards work,
For our RNN, we represent the activation at each node in the AST
by a matrix.  Activations are computed recursively from children
using weights that depend on the type of the node (e.g., whether
it is a for loop or move command), which are then pushed forward
until the root.  Since we use a similar model in the next section
to incorporate style for the previous nonparametric model, we

%We have different parameters for different node types and activations %are matrices. One extra challenge is that the RNN does not have a natural way to express the invocation of user defined methods. Instead, we must unroll method invocations with the method definition. This is highly problematic for methods that take parameters and it impossible for programs that use recursion.

%\Jon{mention dimensions}

While the nonparametric model is more expressive than the RNN formulation, the RNN enjoys a few advantages: (1) It is a more compact representation and can directly share weights across programs and (2) it incorporates the structure of the code, though indirectly. 
 
In practice, we find that the parametric model is more accurate for very simple programs
while the nonparametric model works better for programs of medium complexity and higher.
Though understanding why this is the case largely remains an open question, we believe that
there are at least two factors at play --- (1) the recursive neural network in the parametric
case typically has more layers for more complex programs making it more difficult to train, especially when  and
(2)  the nonparametric programs that we learn approximately obey a
natural property of \emph{composability} while the parametric programs have a much harder
time achieving this property. \Jon{composability needs to be explained.} We explore these differences further in Section~\ref{sec:experiments}.

%While the parametric thing is good for low data
%situations, we hypothesize that it's not so good
%for complex programs.  One reason for this intuition 
%is that it does not capture the notion of composeability
%correctly.  And we can show a plot of 
%composeability of 1 thing, two things, three things..
%
%We can also see on a plot breaking performance up
%into cyclomatic bins that indeed it loses 
%performance on the more complicated programs.
%As a result, while it does well on the simpler HOC
%programs, it does not perform well on the more
%complicated midpoint problem.

\end{versionB}

%#Explain training.  Initialization via auto-encoders for states and %ridge
%regression.  Followed by joint training.
%We can say that we used adagrad.
%Our first goal is to learn an embedding which can capture the %functionality of any program or sub-program.

\vspace{-2mm}
\subsection{Triple Extraction}\label{sec:triplets}
For a given program $S$ we extract Hoare triples by executing it on an exemplar set of unit tests. These tests span a variety of reasonable starting conditions. We instrument the execution of the program such that each time a subtree $A \subset S$ is executed, we record the value, $P$, of all variables before execution, and the value, $Q$, of all variables after execution and save the triple ($P$, $A$, $Q$). We run all programs on unit tests, collecting triples for \emph{all} subtrees. Doing so results in a large dataset $\{(P_i, A_i, Q_i)\}_{i=1}^n$ from which we collapse equivalent
triples.
In practice, some subtrees, especially the body of loops, generate a large (potentially infinite) number of triples. To prevent any subtree from having undue influence on our model we limit the number of triples for any subtree. 

Collecting triples on subtrees, as opposed to just collecting triples on complete programs, is critical since it allows us to learn embeddings not just for the root of a program AST but also for the constituent parts. As a result, we retain data on how a program was implemented, and not just on its overall functionality, which is important for student feedback as we discuss in the next section.
Collecting triples on subtrees also means we are able to optimize our embeddings with substantially more data.
%crucial for both the parametric and non-parametric program embeddings. In the non-parametric setting collecting subtree triples allows us to learn embeddings not just for the root of a program AST but also for the constituent parts.  For the parametric model, collecting triples on subtrees means we are able to train our embeddings with substantially more data. This is especially important as it provides training examples for nodes deeper ASTs with weights that are otherwise hard to learn by recursively back-propagated gradients.
%We share triples across identical subtrees.
\vspace{-2mm}
\section{Feedback Propagation}\label{sec:feedback}
The result of jointly learning to embed states and 
a corpus of programs is a fixed dimensional, real-valued matrix $M_A$ for each subtree $A$ of any program in our corpus. These matrices can be cooperative with machine learning algorithms that can perform tasks beyond predicting what a program does. 
The central application in this paper is the force multiplication of teacher-provided feedback where an active learning algorithm interacts with human graders such that feedback is given to many more assignments than the grader annotates. We propose a two phase interaction. In the first phase, the algorithm selects a subset of exemplar programs for graders to apply a finite set of annotations. Then in the second phase, the algorithm uses the human provided annotations as supervised labels with which it can learn to predict feedback for unlabelled submissions. Each program is annotated with a set $H \subset L$ where $L$ is a discrete collection of $N$ possible annotations. The annotations are meant to cover a range of comments a grader could apply, including feedback on style, strategy and functionality. For each ungraded submission, we must then decide which of the $N$ labels to apply. As such, we view feedback propagation as $N$ binary classification tasks. 

%Ideally, feedback for programs in an educational setting would comment on both functionality and style. Students are learning not just how to write programs that solve tasks, but how to do so in an artful way. 
One way of propagating feedback would be to use the elements of the embedding matrix of the root of a program as features and then train a classifier to predict appropriate feedback for a given program.  However, the matrices we have learned for programs and their subtrees have been trained only to predict functionality.  Consequently, any two programs that are functionally indistinguishable would be given the same instructor feedback under this approach, ignoring any strategic or stylistic differences between the programs.
%This is problematic in the case where the model becomes too capable, %and are able to represent nodes just based on what they do and not the %way in which the code-phrase was implemented. This is clearly the case for the non-parametric program matrix model. 
%Only representing what a program does means that feedback pertaining to style or strategy cannot be predicted.

\subsection{Incorporating structure via
    recursive embedding}\vspace{-1mm}
To recapture the elements of program structure and style 
that are critical for student feedback, our approach to predict feedback uses
the embedding matrices learned for the NPM model, but incorporates
all constituent subtrees of a given AST.
%One could imagine multiple ways of predicting feedback given an abstract syntax tree where each node has a corresponding matrix representation of its observed hoare-triples. We propose that to propagate feedback we learn a model which recursively feeds-forward annotation information up the AST incorporating $M_l$ at each node $l$.
Specifically, using the embedding matrices learned in the NPM model
(which we henceforth denote as $M^{NPM}_A$ for a subtree $A$),
we now propose a new model
based on recursive neural networks (called the \emph{NPM-RNN} model)
in which
we parametrize a matrix $M_A$ in this new model
with an RNN whose
architecture follows the abstract syntax tree 
(similar to the way in which RNN architectures might take the form of a parse tree in an NLP 
setting~\cite{socher2013recursive}).

In our RNN based model, % which we denote the \emph{NPM-RNN model}, 
a subtree of the 
AST rooted at node $j$ is represented
by a matrix which is computed by combining (1) representations of subtrees rooted
at the children of $j$, and (2) the embedding matrix of the subtree
rooted at node $j$ learned via the NPM model.
By incorporating the embedding matrix from the NPM model, we are able to capture the function of every subtree in the AST.

Formally, we will assume each node is associated with some \emph{type}
in set $\mathcal{T}=\{\omega^1, \omega^2,\dots\}$.  Concretely, the type set might
be the collection of keywords or built-in functions that can be called from a program 
in the dataset, e.g., $\mathcal{T}=\{\mathbf{repeat}, \mathbf{while}, \mathbf{if}, \dots\}$.
A node with type $\omega$ is assumed to have a fixed number, $a_\omega$, of children in the AST --- 
for example, a $\mathbf{repeat}$ node has two children, with one child holding the body
of a repeat loop and the second representing the number of times the body is
to be repeated.   

The representation of node $j$ with type $\omega$ 
is then recursively computed in the NPM-RNN model via:\vspace{-3mm}

{\footnotesize
\begin{equation}\label{eqn:feedbackRnn}
a^{(j)} = \phi\left( \sum_{i=1}^{a_\omega} W_i^\omega \cdot a^{(c_i[j])} + b^\omega + \mu M^{NPM}_j \right),
\end{equation}\vspace{-4mm}
}

where: $\phi$ is a nonlinearity (such as $\tanh$), $c_i[j]$ indexes over
the $a_\omega$ children of node $j$, and $M^{NPM}_j$ is the program embedding matrix learned in the NPM model for the subtree rooted at node $j$.  We remind the reader that the activation $a^{(j)}$ at
each node is an $m\times m$ matrix.
Leaf nodes of type $\omega$
are simply associated with a single parameter matrix
$W^\omega$.

In the NPM-RNN model,
we have parameter matrices $W^\omega, b^\omega \in \reals^{m\times m}$ for each possible type 
$\omega \in \mathcal{T}$.
To train the parameters, we
first use the NPM model to compute the 
embedding matrix $M^{NPM}_j$ for each subtree. 
After fixing $M_j$,  we optimize (as with the NPM model)
with minibatch stochastic
gradient descent using backpropagation through structure~\cite{goller1996learning} to compute gradients.
Instead of optimizing for predicting postcondition,
for NPM-RNN, we optimize for each of the binary prediction tasks
that are used for feedback propagation given the vector embedding
at the root of a program. We used hyper-parameters learned in the RNN model optimization since feedback optimization is performed over few examples and without a holdout set.

%Feedback is then predicted using logistic regression off of the %activation of the root, and back-propagated. The parametric model is %already a RNN as such we can simply update the existing parameters.

%\subsection{Active learning}

Finally, feedback propagation has a natural active learning component: 
intelligently selecting submissions for human annotation 
can potentially save instructors significant time.
We find that in practice, running $k$-means on the learned embeddings,
and selecting the cluster centroids as the set of submissions to be annotated works well and leads to significant improvements in feedback propagation
over random subset selection. Surprisingly, having humans annotate the most common programs performs worse than the alternatives, which we
observe to be due to the fact that the most common submissions are all quite similar to one another.

%We experiment with selecting the most popular submissions, randomly %sampled programs, and using clustering. To select exemplar programs %using clustering we run $k$-means on the learned embeddings and select %the most central programs in our dataset.

%We evaluate embeddings both for their ability to propagate feedback given randomly selected graded solutions 
%and when the embeddings are used to choose which partial solutions are graded. 
\vspace{-2mm}
\section{Datasets}\label{sec:data}\vspace{-1mm}

\begin{table}[t]
  \centering
  \ra{1.3}{\footnotesize
  \begin{tabular}{llll}
    \toprule
    
    \tabhead{Statistic} & $\Omega_1$ & $\Omega_1$ & $\Omega_1$  \\
    \midrule
    Num Students & $>$11 million & 2,710 & 2,710\\
    Unique Programs & 210,918 & 6,674 & 63,820\\
    Unique Subtrees & 311,198 & 15,550 & 198,918\\
    Unique Triples & 5,334,452& 476,502 & 4,211,150\\
    Unique States & 149 & 1,399 & 114,704 \\
    Unique Annotations & 15 & 12 & 14 \\
    \bottomrule
  \end{tabular}
  }\vspace{-2mm}
  \caption{Dataset summary. Programs are considered identical if they have equal ASTs. Unique states are different configurations of the gridworld which occur in student programs.}\vspace{-3mm}
  \label{tab:datasummary}
\end{table}

We evaluate our model on three assignments from two different courses, Code.org's Hour of Code (HOC) which has submissions from over 27 million students and Stanford’s Programming Methodology course, a first-term introductory programming course, which has collected submissions over many years from almost three thousand students. From these two classes, we look at three different assignments. As in many introductory programming courses, the first assignments have the students write standard programming control flow (if/else statements, loops, methods) but do not introduce user-defined variables. The programs for these assignments operate in maze worlds where an agent can move, turn, and test for conditions of its current location. In the Stanford assignments, agents can also put down and pick up beepers, 
making the language Turing complete.  Specifically, we study
the following three problems:

$\Omega_1$: The 18$\textsuperscript{th}$ problem in the Hour of Code (HOC). Students solve a task which requires an if/else block inside of a while loop, the most difficult concept in the Hour of Code.

$\Omega_2$: The first assignment in Stanford's course. Students program an agent to retrieve a beeper in a fixed world.

$\Omega_3$: The fourth assignment in Stanford's course. Students program an agent to find the midpoint of a world with unknown dimension. There are multiple  strategies for this problem and many require $O(n^2)$ operations where $n$ is the size of the world. The task is challenging even for those who already know how to program.

In addition to the final submission to any problem, from each student we also collect partial solutions as they progress from starter code to  final answer.  Table~\ref{tab:datasummary} summarizes the sizes of each of the datasets. For all three assignments studied, students take multiple steps to reach their final answer and as a result most programs in our datasets are intermediate solutions that are not responsive to unit tests that simply evaluate correctness. The code.org dataset is available at \url{code.org/research}.

For all assignments we have both functional and stylistic feedback based on  class rubrics which range from observations of solution strategy, to notes on code decomposition, and tests for correctness. The feedback is generated for all submissions (including partial solutions) via a complex script. The script analyzes both the program trees and the series of steps a student took to assign annotations. In general, a script, no matter how complex, does not provide perfect feedback. However the ability to recreate these complex annotations allows us to rigorously evaluate our methods. An algorithm that is able to propagate such feedback should also be able to propagate human quality labels.
\vspace{-1mm}
\section{Results}\label{sec:experiments}\vspace{-1mm}
We rely on a few baselines against which to evaluate our methods,
but the main baseline that we compare to is a simplification of
the NPM-RNN model (which we will call, simply, \emph{RNN}) in which we
drop the program embedding terms $M_j$ from each node 
(cf. Eqn.~\ref{eqn:feedbackRnn}).  

The RNN model can be trained to predict postconditions
as well as to propagate feedback. It has much fewer parameters
than the NPM (and thus NPM-RNN) model
being a strictly parametric model, and is thus expected to 
have an advantage in smaller training set regimes.  On the other hand,
it is also a strictly less expressive model and so the question
is: how much does the expressive power of the NPM and NPM-RNN models
actually help in practice?  We address this question amongst others
using two tasks: predicting postcondition and propagating feedback.

\subsection{Prediction of postcondition}
To understand how much 
  functionality of a program
is captured in our embeddings, we evaluate the accuracy to which we can use the 
program embedding matrices learned by the NPM model
to predict postconditions --- note, however, that
we are not proposing to use the embeddings to predict post-conditions in practice.
We split our observed Hoare triples into training and  test sets and learn our NPM model using the training set. 
Then for each triple $(P, A, Q)$ in the test set we measure how well we can predict the postcondition $Q$ given the corresponding program $A$ and precondition $P$. We evaluate accuracy as the average number of state variables (e.g. row, column, orientation and  location of beepers) that are correctly predicted per triple, and
in addition to the RNN model, compare against the baseline method ``Common" where we select the most common postcondition for a given precondition observed in the training set. 
As our results in Table~\ref{tab:results1} show,
the NPM model achieves the best training accuracy
%is best able to overfit the entire dataset 
(with 98\%, 98\% and 94\% accuracy respectively, for the three problems). 
%and thus represent the observed hoare-triples. 
For the two simpler problems, the parametric (RNN) model achieves slightly better test accuracy, especially for problem $\Omega_2$ where the training set is much smaller. For the most complex programming problem, $\Omega_3$, however, the NPM model substantially outperforms other approaches.

\begin{table}[t]
  \centering
  \begin{tabular}{@{}llll}
     \toprule
        Algorithm & $\Omega_1$ & $\Omega_2$ & $\Omega_3$  \\
    \midrule
    NPM & 95\% (98\%)  & 87\% (98\%) & 81\% (94\%)  \\
    RNN & 96\% (97\%)  & 94\% (95\%) & 46\% (45\%)  \\
    Common & 58\%  &  51\% & 42\% \\
    \bottomrule
  \end{tabular}\vspace{-2mm}
  \caption{\footnotesize Test set postcondition prediction accuracy on 
  the three programming problems. Training set results in parentheses.}\vspace{-4mm}
  \label{tab:results1}
\end{table}

\subsection{Composability of program embeddings}
If we are to represent programs as matrices that act on
a feature space, then a natural desiderata is that they ``compose
well''.  That is, if program $C$ is functionally equivalent
to running program $B$ followed by program $A$, then it should
be the case that $M_C \approx M_B\cdot M_A$.  To evaluate the extent
to which our program embedding matrices are \emph{composable},
we use a corpus of 5000 programs that are composed of a subprogram $A$ followed by another subprogram $B$ (Compose-2). 
%This corpus captures naturally occuring compositions.
We then compare the accuracy of postcondition prediction
using the embedding of an entire program $M_C$ against the
product of embeddings $M_B\cdot M_A$.  
As Table~\ref{tab:results2} shows,
the accuracy using the NPM model for predicting  postcondition is 94\% when using the matrix for the root embedding.  Using the product of two embedding matrices, we see that accuracy does not fall dramatically, with a decoding accuracy of 92\%.
When we test programs that are composed of three subprograms, $A$ followed by $B$, then $C$ (Compose-3), we see 
accuracy drop only to 83\%.

\begin{table}[t]
  \centering
  \begin{tabular}{@{}llllll}
     \toprule
     
         Test & Direct & NPM & NPM-0 & RNN & Common \\
          \midrule
    Compose-2 & 94\% & 92\% & 87\% & 42\% & 39\%\\
Compose-3 & 94\% & 83\% & 72\% & 28\% & 39\%\\
    \bottomrule
  \end{tabular}\vspace{-2mm}
  \caption{\footnotesize Evaluation of composability of  embedding matrices: Accuracy on 5k random triples with ASTs rooted at $\mathbf{block}$ nodes. NPM-0 does not jointly optimize. }\vspace{-4mm}
  \label{tab:results2}
\end{table}

By comparison, the embeddings computed using the 
RNN, a more constrained model, do not seem to satisfy
composability.  We also compare against NPM-0, which is the 
NPM model using just the weights set
by the smart initialization (see Section~\ref{sec:nonparametric}).  While NPM-0 outperforms the RNN,
the full nonparametric model (NPM) performs much better,
suggesting that the joint optimization (of state and program 
embeddings) allows us to learn
an embedding of the state space that is more amenable to composition.

\subsection{Prediction of Feedback}

\begin{figure*}[t]
\centering
\subfigure[]
{
\includegraphics[width=0.3\textwidth]{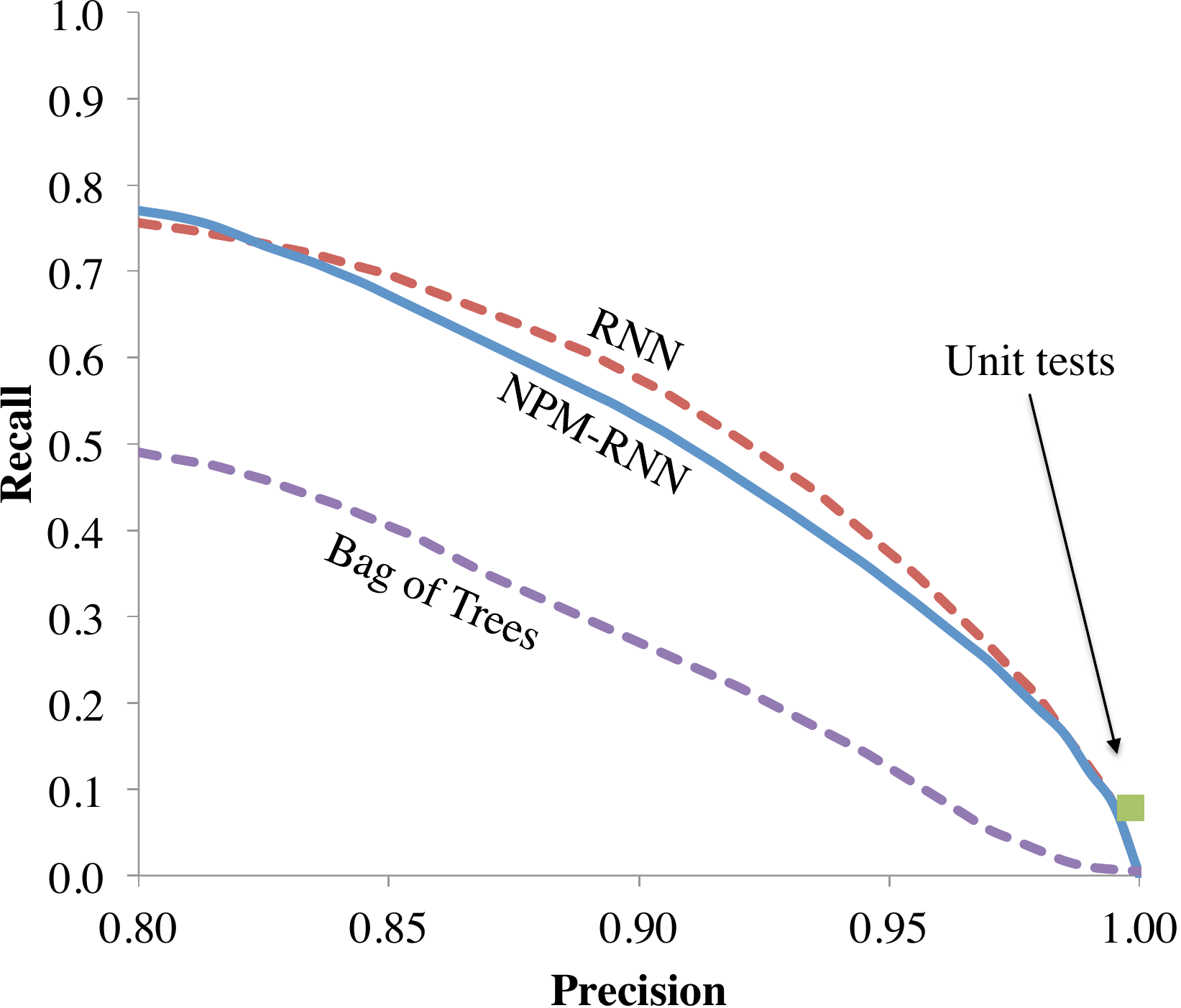}
\label{fig:fmHoc}
}
\subfigure[]
{
\includegraphics[width=0.3\textwidth]{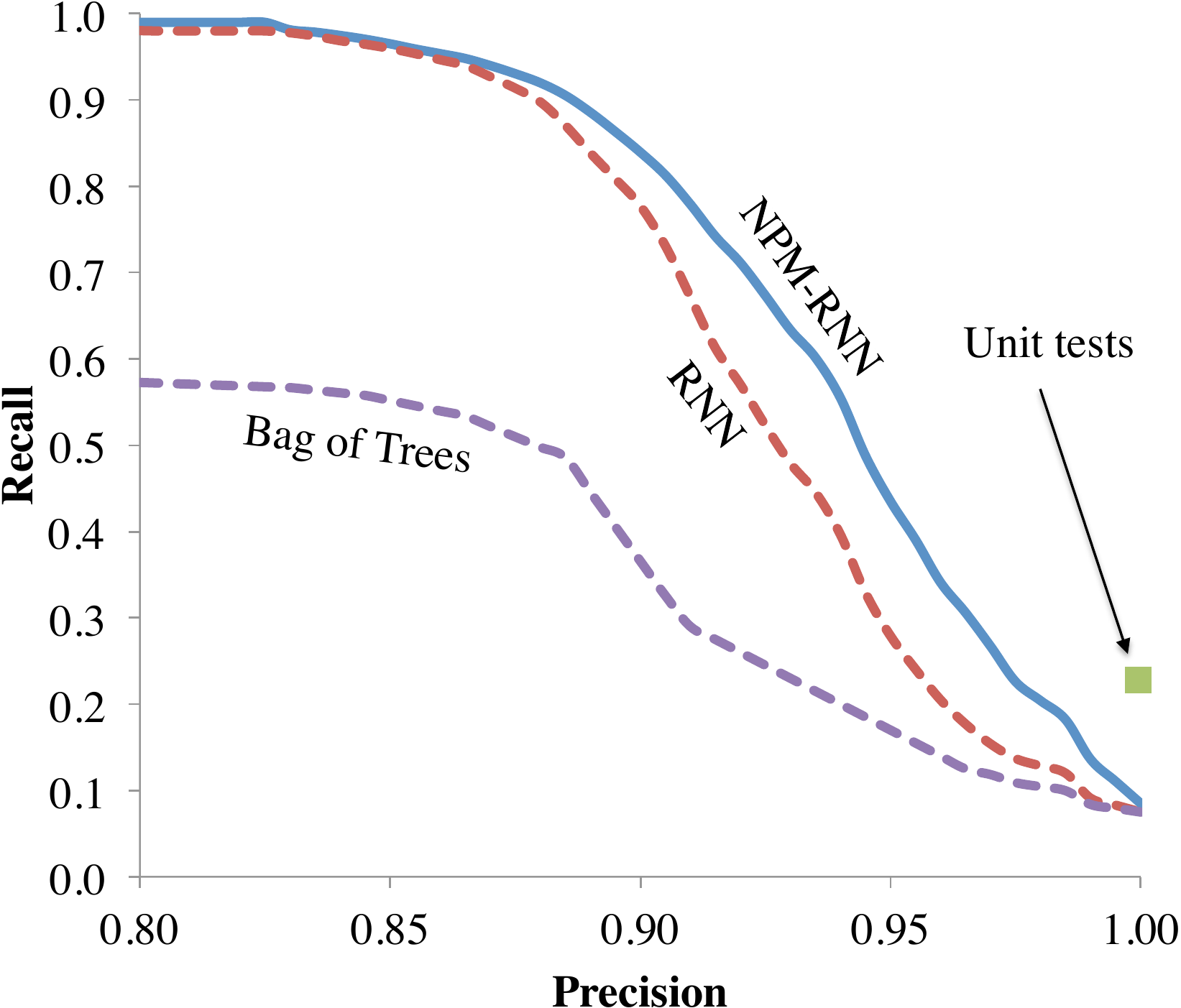}
\label{fig:fmNews}
}
\subfigure[]
{
\includegraphics[width=0.3\textwidth]{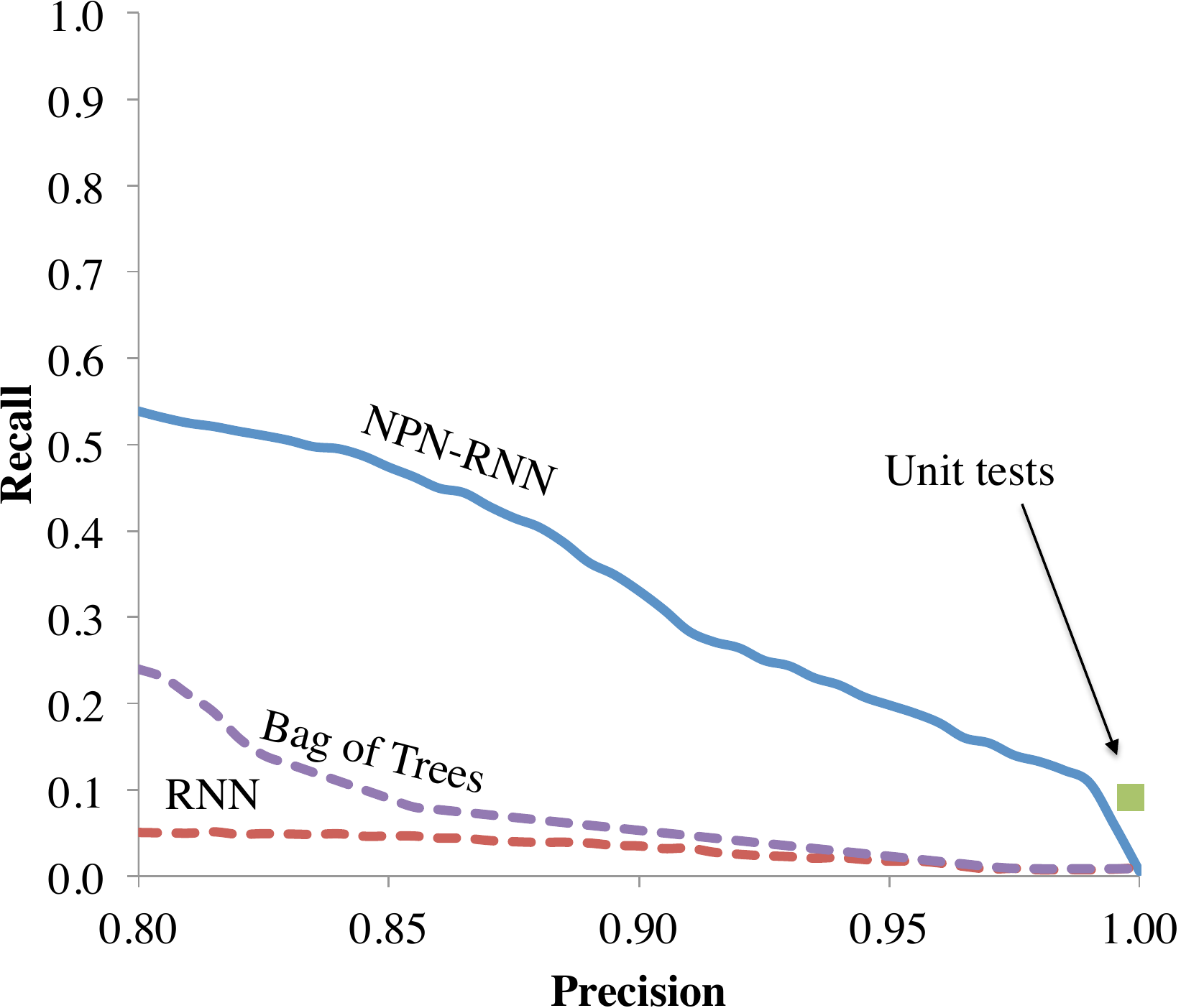}
\label{fig:fmMid}
}\vspace{-4mm}
\caption{\footnotesize
Recall of feedback propagation as a function of precision for three programming problems:
%Interplay between precision and the amount to which we can force multiply feedback for:
\subref{fig:fmHoc} $\Omega_1$,
 \subref{fig:fmNews} $\Omega_2$, and
  \subref{fig:fmMid} $\Omega_3$.  On each, we compare our NPM-RNN against the RNN
  method and two other baselines (bag of trees and unit tests).
 }
\label{fig:force}\vspace{-2mm}
\end{figure*}

We now use our program embedding matrices in the feedback propagation
application described in Section~\ref{sec:feedback}.  
The central question is: given a budget of $K$ human annotated programs (we set $K=500$), what fraction of 
unannotated programs can we propagate these annotations to using the labelled programs, and at what precision?  Alternatively, we are interested in the ``force multiplication factor'' --- the ratio of students who receive feedback via propagation to students to receive human feedback.
%In this experiment, we
%We ran force multiplication with a budget of $K$=500 human annotated programs. We can use the labelled programs and encodings to effectively propagate feedback. 

Figure~\ref{fig:force} visualizes recall and precision of our experiment on each of the three problems. The results translate to 214$\times$, 12$\times$ and 45$\times$ force multiplication factors of teacher effort for $\Omega_1$, $\Omega_2$ and $\Omega_3$ respectively while maintaining 90\% precision. The amount to which we can force multiply feedback depends both on the recall of our model and the size of the corpus to which we are propagating feedback. For example, though $\Omega_2$ had substantially higher recall than $\Omega_1$, in $\Omega_2$ the grading task was much smaller. There were only 6,700 unique programs to propagate feedback to, compared to $\Omega_1$ which had over 210,000. 
As with the previous experiment, we observe that 
for both $\Omega_1$ and $\Omega_2$, the NPM-RNN and RNN models perform similarly. However for $\Omega_3$, the NPM-RNN model substantially outperforms all alternatives.

In addition to the RNN, we compare our results to three other baselines: (1) Running unit tests, (2) a ``Bag-of-Trees'' approach and (3) $k$-nearest neighbor (KNN) with AST edit distances.  The unit tests unsurprisingly are perfect at recognizing correct solutions. However, since our dataset is largely composed of intermediate solutions and not final submissions (especially for $\Omega_1$ and $\Omega_3$), unit tests are not a particularly effective way to propagate annotations. 
%This is best expressed by the distribution of feedback shown in Figure~\ref{fig:whereWrong}, discussed further in the next paragraph.
The Bag-of-Trees approach, where we trained a Na\"{i}ve Bayes model to predict feedback conditioned on the set of subtrees in a program,
is useful
%we found the Naive Bayes probability of feedback, given the %collection of subtrees in a program 
%was notably useful 
for feedback propagation but we observe that it underperforms the embedding solutions on each problem. Moreover, we extended this baseline by amalgamating functionally equivalent code \cite{nguyen14}. Using equivalences found using similar amount of effort as in previous work, we are able to achieve 90\% precision with recall of 39\%, 48\% and 13\%, for the three problems respectively. While this improves the baseline, NPM-RNN obtains almost twice as much recall on all problems. Finally, we find KNN with AST edit distances to be computationally expensive to run and highly ineffective at propagating feedback --- calculating edit distance between all trees requires 20 billion comparisons for $\Omega_1$ and 1.5 billion comparisons for $\Omega_3$. Moreover, the highest precision achieved by KNN for $\Omega_3$ is only 43\% (note that the cut-off for the x-axis in Figure~\ref{fig:force} is 80\%) and at that precision only has a recall of 1.3\%. 

\begin{figure*}[t]
\centering
\subfigure[]
{
\includegraphics[width=0.29\textwidth]{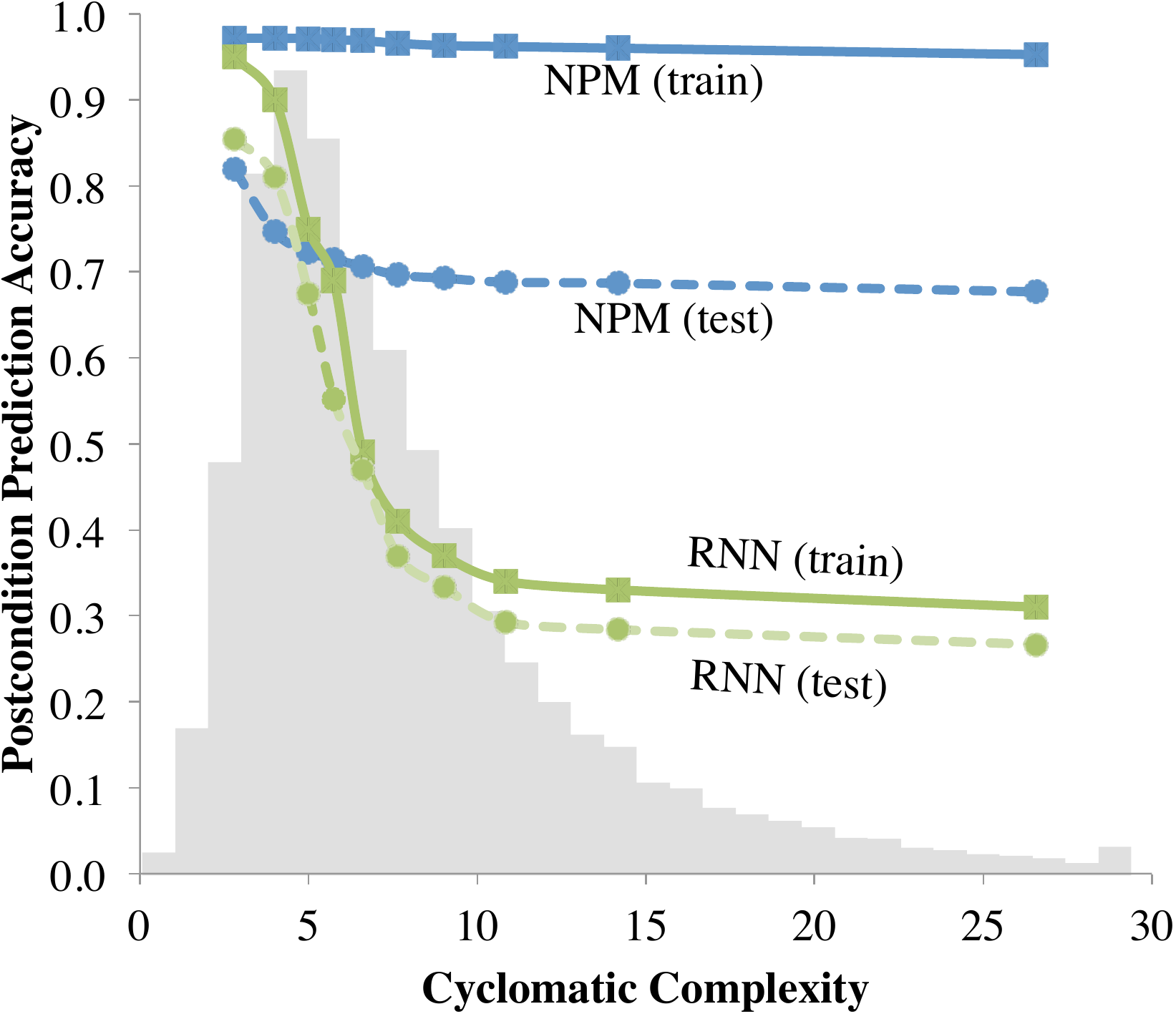}
\label{fig:cycloPost}
}
\subfigure[]
{
\includegraphics[width=0.29\textwidth]{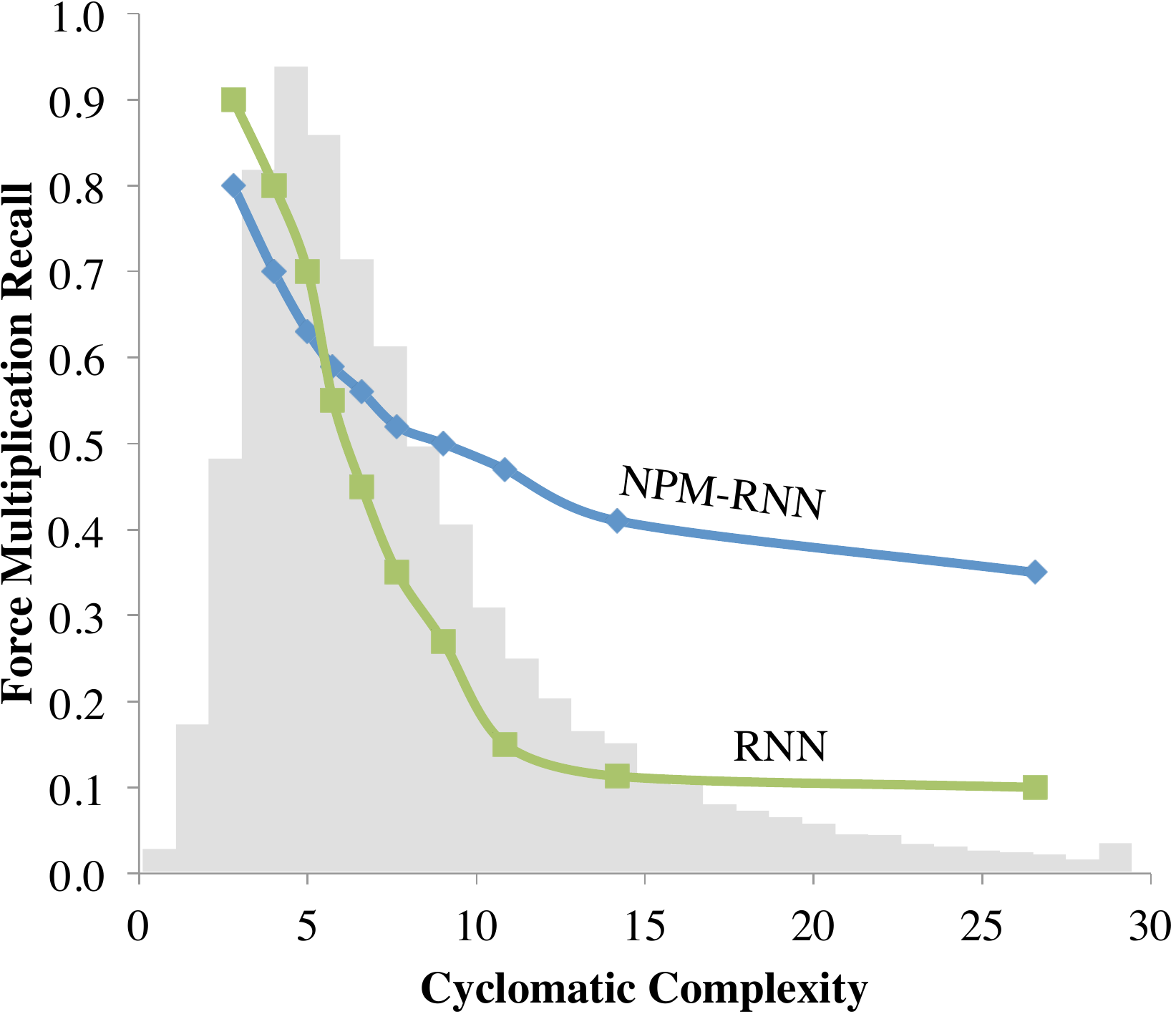}
\label{fig:cycloFm}
}
\subfigure[]
{
\raisebox{4mm}{
\includegraphics[width=0.33\textwidth]{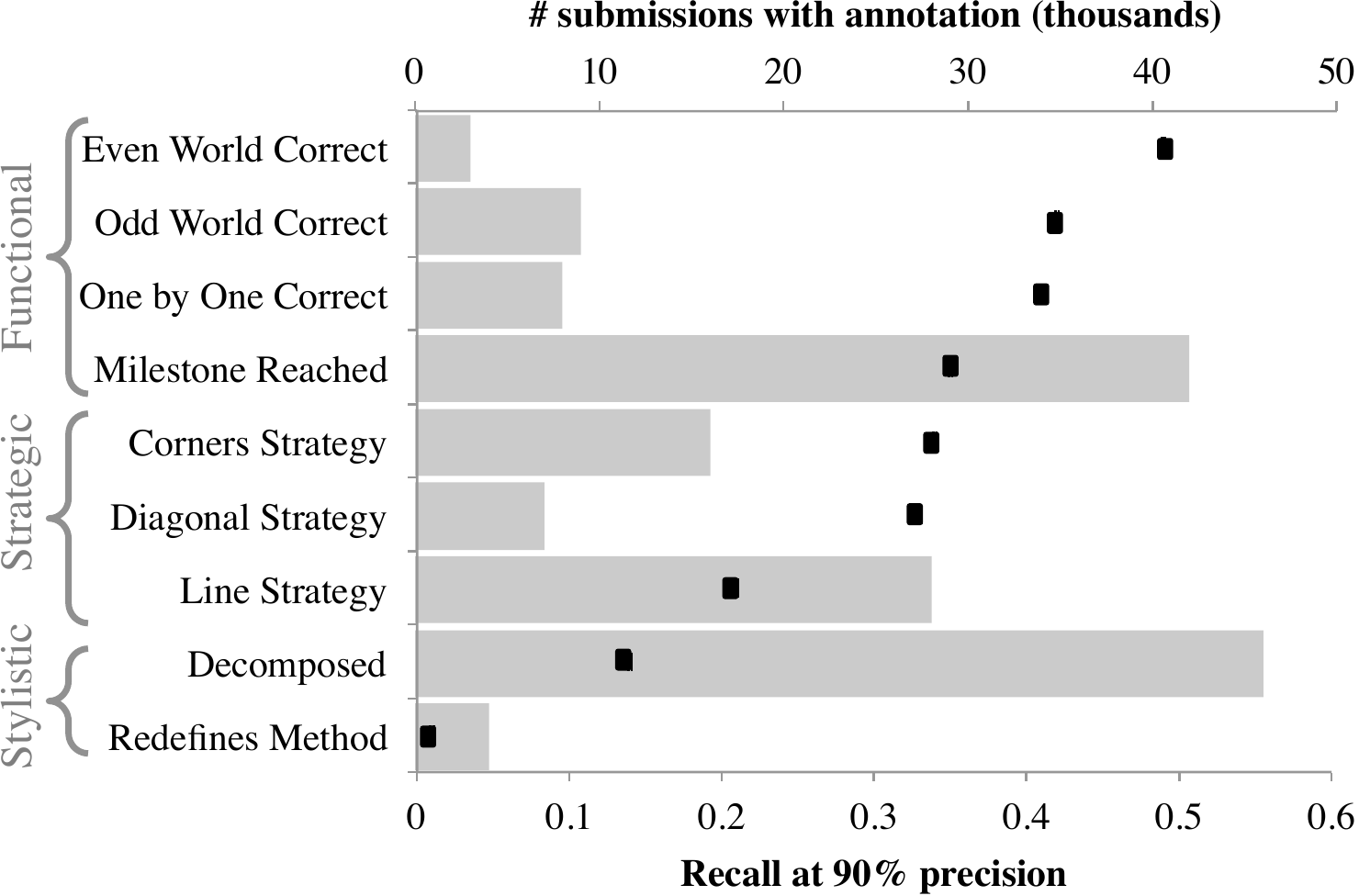}
}
\label{fig:whereWrong}
}\vspace{-4mm}
\caption{\footnotesize
\subref{fig:cycloPost} NPM and RNN postcondition prediction accuracy
as a function of cyclomatic complexity of submitted programs;
\subref{fig:cycloFm}
NPM-RNN and RNN
feedback propagation recall (at 90\% precision). Note that the ratio of human graded assignments to number of programs is much higher in this experiment than Figure~\ref{fig:force}; 
\subref{fig:whereWrong} A breakdown of the accuracy of the nonparametric model
by feedback type for $\Omega_3$ (black dots).  The gray bars histogram
the feedback types by frequency.
}\vspace{-3mm}
\label{fig:cyclomatic}
\end{figure*}

The feedback that we propagate covers a range of stylistic and functional annotations. To further understand the strengths and weaknesses of our solution, we explore the performance of the NPM-RNN model on each of the nine possible annotations for $\Omega_3$. As we see in Figure \ref{fig:whereWrong},
our model performs best on functional feedback with an average 44\% recall at 90\% precision, followed by strategic feedback and performs worst at propagating purely stylistic annotations with averages of 31\% and 8\% respectively.  Overall propagation for $\Omega_3$ is 33\% recall at 90\% precision.

\vspace{-2mm}
\subsection{Code complexity and performance}\vspace{-1mm}
The results from the above experiments are suggestive
that the nonparametric models perform better on more complex
code while the parametric (RNN) model performs better on simpler
code.  To dig deeper, we now look specifically into how our performance depends on the complexity of programs in our corpus
%We set out to better understand the nature of the matrices that we %learned and to do so devised a series of explorations. 
%Our first experiment was to understand how our ability to understand %programs, both predicting post conditions and propagating feedback %depended on the complexity of trees in our corpus. 
 --- a question that is also central to understanding how our models might apply to other assignments. We focus on submissions for $\Omega_3$, which cover a range of complexities, from simple programs to ones with over 50 decision points (loops and if statements). The distribution of \emph{cyclomatic complexity}~\cite{mccabe1976complexity}, a measure of code structure, reflects this wide range (shown in gray in Figures~\ref{fig:cycloPost},\subref{fig:cycloFm}). 
% We sort all submissions to $\Omega_3$ by cyclomatic complexity and bin them into ten groups of equal size. 
We first sort and bin all submissions to $\Omega_3$  by cyclomatic complexity into ten groups of equal size.
Figures~\ref{fig:cycloPost},\subref{fig:cycloFm} plot the results
of the postcondition prediction and force multiplication experiments run individually on these smaller bins (still using a holdout set, and a budget of 500 graded submissions).  
%The most striking result is the differences in performance for prediction of post-condition. 
While the RNN model performs better for simple programs (with cyclomatic complexity $\leq 6$), both train and test accuracies for the RNN degrade dramatically as programs become more complicated. On the other hand, while the NPM model overfits, 
it maintains steady (and better) performance in test accuracy 
as complexity increases. 
This pattern may help to explain our observations that the RNN is more accurate for force multiplying feedback on simple problems.

\vspace{-1mm}
\section{Discussion}\vspace{-1mm}
In this paper we have presented a method for finding simultaneous embeddings of preconditions and postconditions into points in shared Euclidean space where a program can be viewed as a linear 
mapping between these points. 
These embeddings are predictive of the function of a program, and
as we have shown, can be applied to the the tasks of propagating teacher feedback. The courses we evaluate our model on are compelling case studies for different reasons. Tens of millions of students are expected to use Code.org next year,  meaning that the ability to autonomously provide feedback could impact an enormous number of people.  The Stanford course, though much smaller, highlights
the complexity of the code that our method can handle.
%We show the effectiveness of this model applied to the the tasks of propagating teacher feedback. The courses we test our experiment on are compelling case studies for different reasons. Code.org as tens of millions of students
%are expected to use the website next year meaning the ability to autonomously provide feedback could improve the educational experience of many students. The Stanford course, though a high quality curriculum is not taught outside of the university mainly because feedback cannot be delivered in a scaled way. However, while these are worthy applications the programs we analyzed were quite basic. In particular the data we looked at did not include user defined variables which made learning state space more straightforward. To incorporate user defined variables we would need to develop a more sophisticated state model. 

There remains much work towards making these embeddings more 
generally applicable, particularly for domains where we do not have
tens of thousands of submissions per problem or the programs are more complex. For settings where users can define their own variables it would be necessary to find a novel method for mapping program memory into vector space. An interesting
future direction might be to jointly find embeddings across multiple homeworks from the same course, and ultimately, to 
even learn  using arbitrary code outside of a classroom environment. To do so may require more expressive models. From the standpoint  of purely predicting  program output, the approaches described in this paper are not capable of representing arbitrary computation in the 
sense of the Church-Turing thesis.  However,
there has been recent progress in the deep learning community
towards models capable of simulating Turing machines~\cite{graves2014neural}. 
While this  ``Neural Turing Machines'' 
line of work approaches quite a different problem
than our own, we remark that such expressive
representations may indeed be important for statistical
reasoning with arbitrary code databases.
%The utility of embeddings for a corpus of student solutions is not limited to propagating feedback. Having a program representation that is enables general machine learning tasks for example predicting future struggles and even student dropout. Moreover, many courses, including the ones analyzed record not just a student's final solution but also the series of submissions a student took to get there. The task of discovering patterns in these trajectories could be better posed using euclidean embeddings of partial solutions.

For the time being, 
feature embeddings of code can at least be learned using
the massive online education
datasets that have only recently
become available.  And we believe that these features will
be useful in a variety of ways --- not just in propagating 
feedback, but also in tasks such as  predicting future struggles and even student dropout.

%Feature learning holds great promise for assessment of student assignments at scale. For many mediums (eg text, images etc), there are unsupervised learning algorithms that can transform them into a euclidean space, especially given the substantial structure in a corpus of students solving the same problem.

%Other avenues for future work include: joint inference of embedding spaces across assignments, and learning richer models for embeddings and feedback propagation. In this paper we modelled the relationship between Hoare-triples as $f(Q) = M_A \dot f(P)$. While this model was simple and effective one could imagine that it would also be reasonable to embed matrices using $f(Q) = g_A(f(P))$ where $g_A$ is a multi-layered feed-forward network with weights specific to program $A$. It would be harder to interpret the meaning of a program, but the same mapping could be expressed with many fewer parameters.

\section*{Acknowledgments} 
We would like to thank Kevin Murphy, John Mitchell, Vova Kim, Roland Angst, Steve Cooper and Justin Solomon for their critical feedback and useful discussions. We appreciate the generosity of the Code.Org team, especially Nan Li and Ellen Spertus, who providing data and support. Chris is supported by NSF-GRFP grant number DGE-114747.
% also 

\bibliographystyle{icml2013}
\bibliography{paper}

\begin{thebibliography}{22}
\providecommand{\natexlab}[1]{#1}
\providecommand{\url}[1]{\texttt{#1}}
\expandafter\ifx\csname urlstyle\endcsname\relax
  \providecommand{\doi}[1]{doi: #1}\else
  \providecommand{\doi}{doi: \begingroup \urlstyle{rm}\Url}\fi

\bibitem[Basu et~al.(2013)Basu, Jacobs, and Vanderwende]{basu2013powergrading}
Basu, Sumit, Jacobs, Chuck, and Vanderwende, Lucy.
\newblock Powergrading: a clustering approach to amplify human effort for short
  answer grading.
\newblock \emph{Transactions of the Association for Computational Linguistics},
  1:\penalty0 391--402, 2013.

\bibitem[Bergstra \& Bengio(2012)Bergstra and Bengio]{bergstra2012random}
Bergstra, James and Bengio, Yoshua.
\newblock Random search for hyper-parameter optimization.
\newblock \emph{The Journal of Machine Learning Research}, 13\penalty0
  (1):\penalty0 281--305, 2012.

\bibitem[Bowman(2013)]{bowman2013can}
Bowman, Samuel~R.
\newblock Can recursive neural tensor networks learn logical reasoning?
\newblock \emph{arXiv preprint arXiv:1312.6192}, 2013.

\bibitem[Brooks et~al.(2014)Brooks, Basu, Jacobs, and
  Vanderwende]{brooks2014divide}
Brooks, Michael, Basu, Sumit, Jacobs, Charles, and Vanderwende, Lucy.
\newblock Divide and correct: Using clusters to grade short answers at scale.
\newblock In \emph{Proceedings of the first ACM conference on Learning@ scale
  conference}, pp.\  89--98. ACM, 2014.

\bibitem[Duchi et~al.(2011)Duchi, Hazan, and Singer]{duchi2011adaptive}
Duchi, John, Hazan, Elad, and Singer, Yoram.
\newblock Adaptive subgradient methods for online learning and stochastic
  optimization.
\newblock \emph{The Journal of Machine Learning Research}, 12:\penalty0
  2121--2159, 2011.

\bibitem[Goller \& Kuchler(1996)Goller and Kuchler]{goller1996learning}
Goller, Christoph and Kuchler, Andreas.
\newblock Learning task-dependent distributed representations by
  backpropagation through structure.
\newblock In \emph{Neural Networks, 1996., IEEE International Conference on},
  volume~1, pp.\  347--352. IEEE, 1996.

\bibitem[Graves et~al.(2014)Graves, Wayne, and Danihelka]{graves2014neural}
Graves, Alex, Wayne, Greg, and Danihelka, Ivo.
\newblock Neural turing machines.
\newblock \emph{arXiv preprint arXiv:1410.5401}, 2014.

\bibitem[Hoare(1969)]{hoare1969axiomatic}
Hoare, Charles Antony~Richard.
\newblock An axiomatic basis for computer programming.
\newblock \emph{Communications of the ACM}, 12\penalty0 (10):\penalty0
  576--580, 1969.

\bibitem[Huang et~al.(2013)Huang, Piech, Nguyen, and Guibas]{huangetal13}
Huang, Jonathan, Piech, Chris, Nguyen, Andy, and Guibas, Leonidas~J.
\newblock Syntactic and functional variability of a million code submissions in
  a machine learning mooc.
\newblock In \emph{The 16th International Conference on Artificial Intelligence
  in Education (AIED 2013) Workshop on Massive Open Online Courses (MOOCshop)},
  2013.

\bibitem[Lan et~al.(2015)Lan, Vats, Waters, and Baraniuk]{lan2015mathematical}
Lan, Andrew~S, Vats, Divyanshu, Waters, Andrew~E, and Baraniuk, Richard~G.
\newblock Mathematical language processing: Automatic grading and feedback for
  open response mathematical questions.
\newblock \emph{arXiv preprint arXiv:1501.04346}, 2015.

\bibitem[McCabe(1976)]{mccabe1976complexity}
McCabe, Thomas~J.
\newblock A complexity measure.
\newblock \emph{Software Engineering, IEEE Transactions on}, \penalty0
  (4):\penalty0 308--320, 1976.

\bibitem[Mokbel et~al.(2013)Mokbel, Gross, Paassen, Pinkwart, and
  Hammer]{mokbel2013domain}
Mokbel, Bassam, Gross, Sebastian, Paassen, Benjamin, Pinkwart, Niels, and
  Hammer, Barbara.
\newblock Domain-independent proximity measures in intelligent tutoring
  systems.
\newblock In \emph{Proceedings of the 6th International Conference on
  Educational Data Mining (EDM)}, 2013.

\bibitem[Nguyen et~al.(2014)Nguyen, Piech, Huang, and Guibas]{nguyen14}
Nguyen, Andy, Piech, Christopher, Huang, Jonathan, and Guibas, Leonidas.
\newblock Codewebs: Scalable homework search for massive open online
  programming courses.
\newblock In \emph{Proceedings of the 23rd International World Wide Web
  Conference (WWW 2014)}, Seoul, Korea, 2014.

\bibitem[Ovsjanikov et~al.(2012)Ovsjanikov, Ben-Chen, Solomon, Butscher, and
  Guibas]{ovsjanikov2012functional}
Ovsjanikov, Maks, Ben-Chen, Mirela, Solomon, Justin, Butscher, Adrian, and
  Guibas, Leonidas.
\newblock Functional maps: a flexible representation of maps between shapes.
\newblock \emph{ACM Transactions on Graphics (TOG)}, 31\penalty0 (4):\penalty0
  30, 2012.

\bibitem[Ovsjanikov et~al.(2013)Ovsjanikov, Ben-Chen, Chazal, and
  Guibas]{ovsjanikov2013analysis}
Ovsjanikov, Maks, Ben-Chen, Mirela, Chazal, Frederic, and Guibas, Leonidas.
\newblock Analysis and visualization of maps between shapes.
\newblock In \emph{Computer Graphics Forum}, volume~32, pp.\  135--145. Wiley
  Online Library, 2013.

\bibitem[Piech et~al.(2015)Piech, Sahami, Huang, and Guibas]{piech2015}
Piech, Chris, Sahami, Mehran, Huang, Jonathan, and Guibas, Leonidas.
\newblock Autonomously generating hints by inferring problem solving policies.
\newblock In \emph{Proceedings of the Second (2015) ACM Conference on Learning
  @ Scale}, L@S '15, pp.\  195--204. ACM, 2015.

\bibitem[Rogers et~al.(2014)Rogers, Garcia, Canny, Tang, and
  Kang]{rogers2014aces}
Rogers, Stephanie, Garcia, Dan, Canny, John~F, Tang, Steven, and Kang, Daniel.
\newblock \emph{ACES: Automatic evaluation of coding style}.
\newblock PhD thesis, Master’s thesis, EECS Department, University of
  California, Berkeley, 2014.

\bibitem[Socher et~al.(2011)Socher, Pennington, Huang, Ng, and
  Manning]{socher2011semi}
Socher, Richard, Pennington, Jeffrey, Huang, Eric~H, Ng, Andrew~Y, and Manning,
  Christopher~D.
\newblock Semi-supervised recursive autoencoders for predicting sentiment
  distributions.
\newblock In \emph{Proceedings of the Conference on Empirical Methods in
  Natural Language Processing}, pp.\  151--161. Association for Computational
  Linguistics, 2011.

\bibitem[Socher et~al.(2013)Socher, Perelygin, Wu, Chuang, Manning, Ng, and
  Potts]{socher2013recursive}
Socher, Richard, Perelygin, Alex, Wu, Jean~Y, Chuang, Jason, Manning,
  Christopher~D, Ng, Andrew~Y, and Potts, Christopher.
\newblock Recursive deep models for semantic compositionality over a sentiment
  treebank.
\newblock In \emph{Proceedings of the Conference on Empirical Methods in
  Natural Language Processing (EMNLP)}, pp.\  1631--1642. Citeseer, 2013.

\bibitem[Song et~al.(2009)Song, Huang, Smola, and Fukumizu]{song2009hilbert}
Song, Le, Huang, Jonathan, Smola, Alex, and Fukumizu, Kenji.
\newblock Hilbert space embeddings of conditional distributions with
  applications to dynamical systems.
\newblock In \emph{Proceedings of the 26th Annual International Conference on
  Machine Learning}, pp.\  961--968. ACM, 2009.

\bibitem[Song et~al.(2013)Song, Fukumizu, and Gretton]{song2013kernel}
Song, Le, Fukumizu, Kenji, and Gretton, Arthur.
\newblock Kernel embeddings of conditional distributions: A unified kernel
  framework for nonparametric inference in graphical models.
\newblock \emph{Signal Processing Magazine, IEEE}, 30\penalty0 (4):\penalty0
  98--111, 2013.

\bibitem[Zaremba et~al.(2014)Zaremba, Kurach, and Fergus]{zaremba2014learningb}
Zaremba, Wojciech, Kurach, Karol, and Fergus, Rob.
\newblock Learning to discover efficient mathematical identities.
\newblock In \emph{Advances in Neural Information Processing Systems}, pp.\
  1278--1286, 2014.

\end{thebibliography}

\end{document}